\documentclass[11pt,a4paper]{article}
\usepackage{microtype}
\usepackage[utf8]{inputenc}
\usepackage[small,bf]{caption}
\usepackage[english]{babel}
\usepackage{graphicx}
\usepackage{url}
\usepackage{multirow}
\usepackage{natbib}

\title{Using external sources of bilingual information for on-the-fly word alignment}

 \author{Miquel Esplà-Gomis, Felipe Sánchez-Martínez and Mikel L.\ Forcada \\
   \emph{Departament de Llenguatges i Sistemes Informàtics} \\
 		\emph{Universitat d'Alacant, E-03071 Alacant, Spain} \\
   \texttt{mespla@dlsi.ua.es, fsanchez@dlsi.ua.es, mlf@dlsi.ua.es} \\
 }

\date{Technical report, December 7, 2012}

\begin{document}

\maketitle

%
\begin{abstract}
  In this paper we present a new and simple language-independent
  method for word-alignment based on the use of external sources of
  bilingual information such as machine translation systems. We show
  that the few parameters of the aligner can be trained on a very
  small corpus, which leads to results comparable to those obtained by
  the state-of-the-art tool GIZA++ in terms of precision. Regarding
  other metrics, such as alignment error rate or \(F\)-measure, the
  parametric aligner, when trained on a very small gold-standard (450
  pairs of sentences), provides results comparable to those produced
  by GIZA++ when trained on an in-domain corpus of around 10,000 pairs
  of sentences.  Furthermore, the results obtained indicate that the
  training is domain-independent, which enables the use of the trained
  aligner \emph{on the fly} on any new pair of sentences.
%
\end{abstract}

\section{Introduction}

\subsection{The need for word [position] alignment}

Corpus-based translation technologies use information obtained from
existing \emph{segment pairs}, that is, pairs of text segments which
are a translation of each other ---such as (\emph{Give the book to
  me}, \emph{Donne-moi le livre})---, to perform a translation
task. These pairs of segments are usually, but not always,
\emph{sentence} pairs, and to be able to translate new, unseen text
segments, the information in them is usually generalized after
performing \emph{word alignment}. The task of word alignment consists
in determining the correspondence between the words (actually word
positions) in one segment and those in the other segment. After word
alignment, smaller sub-segment \emph{translation units}, such as
(\emph{le livre}, \emph{the book}), can be extracted. These
translation units have a prominent role in state-of-the-art
statistical machine translation (SMT, \citep{koehn2009}), and are
usually referred to as \emph{phrase pairs} in the SMT literature.

The most widely used alignment method is based on the so-called IBM
models by \citet{brown93} and the HMM-based alignment model by
\citet{vogel96}, both implemented in the free/open-source GIZA++ tool
\citep{och03asc}.\footnote{\url{http://code.google.com/p/giza-pp/}[last
  visit: 30th August 2012]} Roughly, these methods, which were devised
for building word-based SMT systems, establish correspondences between
the word positions in one segment and the word positions in the other
segment of the pair by using iterative expectation-maximization (EM)
training on large sets of segment pairs called \emph{parallel corpora}
(also \emph{translation memories} in computer-aided translation, CAT).

The two key components of the EM approach to word alignment are: (a)
the building of probabilistic dictionaries that model the
correspondence between the words (not word positions) in one language
and those in the other language, independently of the actual segment
pairs in which they were found; and (b) the building of rather
sophisticated statistical \emph{alignment models} which explicitly
model \emph{fertility} (the maximum number of words with which a word
can be aligned) and \emph{reorderings}, and that use the probabilistic
dictionaries to describe the alignment in each segment pair.
EM iterations improve these two probabilistic models alternatively by
approximately assigning an increasing likelihood to the training
corpus in each iteration; the quality of the estimation and the
training time both increase with the size of the parallel corpus
(roughly linearly, \citep{toral2012p}). The resulting probability
models are then used to extract the best word-position alignment,
usually called just \emph{word alignment}, in each sentence pair.

\subsection{The need for \emph{on-the-fly} word [position] alignment}

While the state-of-the-art approach to word alignment is appropriate
as a first step when building an SMT system, it may happen to be
unfeasible because the parallel corpus available is not large enough
to get accurate word alignments, or because it is too costly in terms
of time. This is actually the case when one needs to \emph{word-align}
a few new segment pairs \emph{on the fly}, that is, instantaneously,
for instance, when performing CAT using translation memories, as in
the case of the works by \citet{kranias2004automatic} and
\citet{espla11b}.\footnote{For the use of word-position alignment
  information in CAT, see \citet{espla11b} and
  \citet{kuhn2011patent}.} There is, of course, the possibility of
using a probabilistic alignment model previously trained on another,
ideally related, parallel corpus to align the word positions in the
new segment pairs; however, these pre-trained alignment models may not
be generally available for every possible domain or task.

We describe alternative ways to perform \emph{word-position} alignment
on a segment pair, on the fly and on demand, by using readily
available sources of translation units, which we will refer to as
\emph{sources of bilingual information} (SBI); for instance, existing
(on-line) machine translation systems. Information from the SBI is
initially used to discover correspondences between variable-length
sub-segments in the pair of segments to align, and then processed to
obtain word-position alignments. The word-position alignments are
obtained by applying a probabilistic word-position model whose
parameters have to be trained on a parallel corpus; no assumptions are
made about the pair of languages involved. The corpus, as it will be
shown, need not be related to the new segment pairs being word
aligned; parameters are therefore transferable across text domains.
In addition, there is a particular choice of parameters that
completely avoids the need for training and has an intuitive
``physical'' interpretation, yielding reasonably good results.


\subsection{Related work}

In addition to the IBM models and the HMM alignment model previously
mentioned, one can find in the literature different approaches to the
problem of word-position alignment. In this section we focus on those
approaches that make use of SBI in some way; for a complete review of
the state of the art in word alignment the reader is referred to
\citet{tiedemann2011bitext}.



\citet{Fung1997} introduces the use of a bilingual dictionary as a SBI
to obtain an initial alignment between \emph{seed words} in a parallel
corpus. These seed words are chosen so that they cannot have multiple
translations (in both languages) and are frequent enough to become
useful references in both texts of the parallel corpus. These initial
alignments are then used to align the other words appearing around
them in the parallel texts using an heuristic method similar to the
one introduced by \citet{Rapp:1999:AIW:1034678.1034756}.

\citet{liu2005log} propose the use of a \emph{log-linear}
(maximum-entropy style) model \citep{berger96} to combine the IBM
model 3 alignment model with information coming from part-of-speech
taggers and bilingual dictionaries; the work was later extended to
include new features and a new training procedure
\citep{liu2010discriminative}. The main differences between their work
and the one presented here are: (i) we do not rely on any previously
computed alignment model; (ii) we use any possible SBI which may
relate multi-word segments, and (iii) they model the word-position
alignment task as a \emph{structured prediction problem}
\citep[p. 82]{tiedemann2011bitext} that generates the whole alignment
structure, whereas we model each association of positions
independently. We will further discuss this last difference in the
next section.

 
\section{The alignment model} \label{se:alg-model}

The method we present here uses the available sources of bilingual
information (SBI) to detect parallel sub-segments in a given pair of
parallel text segments \(S\) and \(T\) written in different
languages. Once sub-segment alignments have been identified, the
word-position alignments are obtained after computing the probability
\(p(j,k)\) of every pair of word positions \((j,k)\) being
aligned. For the computation on these probabilities a set of feature
functions are used which are based on the sub-segment alignments
observed.

%

We define the probability \(p(j,k)\) as follows:
\begin{equation}
  \label{eq:prob}
  p(j,k) = \displaystyle \exp\left(\sum_{p=1}^{n_F}\lambda_p
      f_p(j,k)\right) \left(\displaystyle  \sum_{k'}\sum_{j'}\exp\left(\sum_{p=1}^{n_F} \lambda_p f_p(j',k')\right)\right)^{-1}
\end{equation}
where (a) the source-side position indexes \(j\) (also \(j'\)) can
take values from 1 to \(|S|\), but also be NULL, and target-side
position indexes \(k\) (also \(k'\)) can take values from 1 to
\(|T|\), and also be NULL, but never simultaneously to a source-side
index (alignments from NULL to NULL are not possible); and (b)
\(f_p(j,k)\) is the $p$-th feature (see below) relating the \(j\)-th
word of the source sentence \(S\) and the \(k\)-th word of the target
sentence \(T\). This is a maximum-entropy-style function that is
always in \([0,1]\) and that has the property that
\begin{displaymath}
\sum_{k}\sum_{j} p(j,k) = 1
\end{displaymath}
when summing for all valid index pairs.  The probabilities \(p(j,k)\)
may be interpreted as the probability that someone who does not know
the languages involved links position \(j\) in \(S\) and \(k\) in
\(T\) after looking at the set of translation pairs provided by the
SBI which happen to match sub-segments in \(S\) and \(T\).

This model is similar to the one proposed by \citet{liu2005log} and
later by \citet{liu2010discriminative} as discussed in the previous
section. One important difference between both models is that these authors
formulate the alignment as a \emph{structured prediction problem} in
which the probability for a pair of segments is computed for the whole
set of word-position alignments \(a=\{(j,k)\}\); that is, the
probability of a word-position alignment \((j,k)\) gets influenced by
the rest of word-positions alignments for that pair of segments. In
contrast, we model each word-position alignment independently. This
may be less expressive but has interesting advantages from the
computational point of view when searching for the best set of
word-position alignments for a pair of segments.


\paragraph{Sub-segment alignment.}
To obtain the sub-segment alignments, both segments \(S\) and \(T\)
are segmented in all possible ways to obtain sub-segments of length
\(l \in [1,L]\), where \(L\) is a given maximum sub-segment length
measured in words.  Let \(\sigma\) be a sub-segment from \(S\) and
\(\tau\) a sub-segment from \(T\). We consider that \(\sigma\) and
\(\tau\) are aligned if any of the available SBI confirm that
\(\sigma\) is a translation of \(\tau\), or vice versa.

\begin{figure}[!ht]
  \centering
  \includegraphics[scale=0.6]{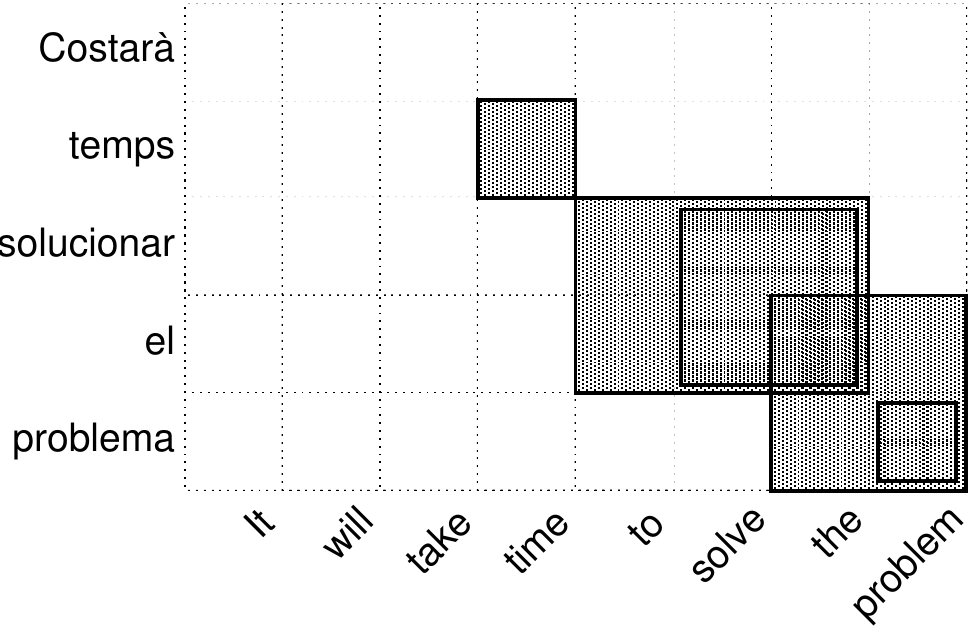}
  \caption{Sub-segment alignments.}
  \label{fig:nalignments}
\end{figure}

Suppose the pair of parallel segments \(S\)=\emph{Costarà temps
  solucionar el problema}, in Catalan, and \(T\)=\emph{It will take
  time to solve the problem}, in English. We first obtain all the
possible sub-segments \(\sigma\) in \(S\) and \(\tau\) in \(T\) and
then use machine translation as a SBI by translating the sub-segments
in both translation directions. We obtain the following set of
sub-segment alignments:
\begin{center}
\small{
\begin{tabular}{rcl}
 \emph{temps} &\(\leftrightarrow\)& \emph{time} \\
 \emph{problema} &\(\leftrightarrow\)& \emph{problem} \\
 \emph{solucionar el} &\(\to\)& \emph{solve the} \\
 \emph{solucionar el} &\(\gets\)& \emph{to solve the} \\
 \emph{el problema} &\(\leftrightarrow\)& \emph{the problem} \\
\end{tabular}
}
\end{center}
It is worth noting that multiple alignments for a sub-segment are
possible, as in the case of the sub-segment \emph{solucionar el} which
is both aligned with \emph{solve the} and \emph{to solve the}.  In
those cases, all the sub-segment alignments available are used.
Figure \ref{fig:nalignments} shows a graphical representation of these
alignments.

\paragraph{Features.} 
The information provided by the sub-segment alignments is used to
build the features that are combined to compute the probabilities
\(p(j,k)\) through eq.~(\ref{eq:prob}). This feature functions
are based on the function
\(\mathrm{cover}(j,k,\sigma,\tau)\),
which equals 1 if sub-segment \(\sigma\) \emph{covers} the \(j\)-th
word in \(S\) and \(\tau\) \emph{covers} the \(k\)-th word in \(T\),
and 0 otherwise.
In particular, by considering sub-segments \(\sigma\) and \(\tau\) of
lengths \(m\) and \(n\) varying from 1 to the maximum sub-segment
length \(L\) we define the following set of \(L^2\) features, one
feature for each possible combination of lengths \((m,n) \in [1,L]
\times [1,L]\):
\begin{displaymath}
f_{(m-1)L+n}=\sum_{(\sigma,\tau)\in M(S,T), |\sigma|=m, |\tau|=n} \mathrm{cover}(j,k,\sigma,\tau),
\end{displaymath}
where \(|x|\) stands for the length of sub-segment \(x\) measured in
words.\footnote{One may also split this feature set to treat each
  different SBI separately or even lift the restriction on the source
  and target lengths \(m\) and \(n\), and build new features depending
  only on \(n\) and \(m\), respectively.}

\paragraph{Alignment computation.}

To get the word-position alignments of a pair of segments \(S\) and
\(T\) we follow a greedy method that makes two simplifying
assumptions:
\begin{itemize}\itemsep 0ex
\item each word position \(j\) in \(S\) is aligned to either a single
  word position \(k\) in \(T\) or to NULL (source-to-target
  alignment);
\item then, independently, each word position \(k\) in \(T\) is
  aligned to either a single word position \(j\) in \(S\) or to NULL
  (target-to-source alignment).
\end{itemize}
Therefore all possible alignments of sentences \(S\) and \(T\) have
exactly \(|S|+|T|\) alignments. The total probability of each such
alignment \(a\) is
\begin{equation}
p(a) = \prod_{(j,k)\in a} p(j,k) = \prod_{j=1}^{|S|} p(j,k^\star(j)) \times
\prod_{k=1}^{|T|} p(j^\star(k),k),
\end{equation} 
where each position \(j\) in \([1,|S|]\) aligns to a single position \(k^\star(j)\) in \([1,|T|]\cup\{\mathrm{NULL}\}\), and each position \(k\) in \([1,|T|]\) aligns to a single position \(j^\star(k)\) in \([1,|S|]\cup\{\mathrm{NULL}\}\).
It may be easily shown that if we choose
\begin{equation}
j^\star(k)=\left\{
\begin{array}{cl} 
\mathrm{arg}\max_{1\le j\le|S|} p(j,k) & \mbox{if } p(j,k)>1/Z \\
\mathrm{NULL} & \mbox{otherwise} 
\end{array}
\right.
\end{equation}
and
\begin{equation}
k^\star(j)=\left\{\begin{array}{cl} \mathrm{arg}\max_{1\le k\le|T|} p(j,k) & \mbox{if } p(j,k)>1/Z \\
\mathrm{NULL} & \mbox{otherwise} .\end{array}
\right.
\end{equation}
the resulting alignment probability is the highest possible. The case
\(p(j,k)=1/Z\) where \(Z\) is the normalizing factor on the right side
of eq.~(\ref{eq:prob}) occurs when no evidence has been found for that
particular position pair \((j,k)\),
i.e. \(\textrm{cover}(j,k,\sigma,\tau)\) is zero; in that case, we
decide to align these words to NULL. In case of finding two
equiprobable alignment candidates for a given word, the one closest to
the diagonal is chosen.

Note that the above alignments may be considered as two separate sets
of asymmetrical alignments that may be symmetrized as is usually done
with statistical alignments. The union alignment is the whole set of
\(|S|+|T|\) alignments; the intersection and
\emph{grow-diagonal-final-and} \citep{koehn03statis} alignments can
also be readily obtained from them.

\paragraph{Training.}  
%
To get the best values of \(\lambda_p\) we try to fit our alignments 
to the reference alignments \(\hat{a}_m\) in a training corpus \(C\) of \(n_S\) sentences. We do this in
basically two ways. 

The first one consists in \textbf{maximizing the probability} (actually the logarithm of the probability) 
of the whole training corpus \(C\):
\begin{equation}
  \label{eq:maximizeprobability}
  \log p(C) = \sum_{m=1}^{n_S} \sum_{(j,k)\in \hat{a}_m} \log p(j,k;m)
\end{equation}
where indexes \(j\) and \(k\) can be NULL as explained above (unaligned words
in the reference alignment \(\hat{a}_m\) are assumed to be aligned to
NULL). Sentence index \(m\) has been added to the probability function for clarity.
Eq.~(\ref{eq:maximizeprobability}) is differentiable with respect to the parameters \(\lambda_p\), which allows for gradient ascent training, with each component of the gradient computed as follows:
\begin{equation}
   \label{eq:deriv}
\frac{\partial E}{\partial \lambda_p}= \sum_{m=1}^{n_S}  \sum_{(j,k)\in \hat{a}_m}
\left( f_p(j,k;m)-\sum_{(j',k')\in \hat{a}_m} p(j',k';m)f_p(j',k';m) \right),
\end{equation}
where sentence index \(m\) has been also added to \(f_p(j,k)\) for the
sake of clarity.

The second approach tries to \textbf{minimize directly an alignment
  error measure} that indicates how much a discretized, symmetrized
alignment obtained by our method departs from the alignments observed
in the training corpus: for instance, the alignment error rate
(AER)~\citep{och03asc} or \(1-F\) where \(F\) is the \(F\)-measure
\citep[Ch.\ 8.1]{manning1999foundations},
much as it is done by \citep{liu2010discriminative}. Discretization
renders these error measures non-differentiable; therefore, we resort
to using general-purpose function optimization methods such as the
multidimensional simplex optimization of
\citep{nelder1965simplex}.\footnote{\protect\citet{liu2010discriminative}
  use MERT instead.}





With the two approaches the number of trainable parameters is small
(of the order of \(L^2\), where \(L\) is the maximum sub-segment
length considered), therefore reasonable results may be expected with
a rather small training corpus and a SBI covering well the sentence
pairs. This is because no word probabilities have to be learned but
only parameters to produce word-position alignments using information
from the SBIs.

\subsection{An intuitive aligner that does not need training}
\label{ss:pressure}

There is a set of parameters for the model described above that has an
intuitive ``physical'' interpretation, and that yields reasonable
results, as shown in Section~\ref{se:exp}. This set of parameters
could be used as a starting point for optimization or as a first
approximation.  

If one chooses \(\lambda_{(m-1)L+n} = ({mn})^{-1}\),
eq.~(\ref{eq:prob}) may be rewritten as:
\begin{displaymath}
p(j,k)=\exp(P_{jk}(S,T,M(S,T))\left(\sum_{j'}\sum_{k'}\exp(P_{j'k'}(S,T,M(S,T))\right)^{-1}
\end{displaymath}
where the \emph{alignment presssure} \(P_{jk}(S,T,M(S,T))\) between the
\(j\)-th word in \(S\) and the \(k\)-th word in \(T\) is
\begin{eqnarray*}
  P_{jk}(S,T,M(S,T))=\sum_{(\sigma,\tau)\in M(S,T)} \frac{\mathrm{cover}(j,k,\sigma,\tau)}{|\sigma| \cdot |\tau|}
\end{eqnarray*}
where $M(S,T)$ is the set of sub-segment alignments detected for the
pair of parallel segments \(S\) and \(T\). If either \(j\) or \(k\) are NULL,
\(\mathrm{cover}(j,k,\sigma,\tau)\) is zero.

%

Intuitively, each \(P_{jk}\) may be seen as the \emph{pressure}
applied by the sub-segment alignments on the word pair \((j,k)\); so
the wider the surface (\(|\sigma||\tau|\)) covered by a sub-segment alignment, the lower the contribution of that sub-segment
pair to the total pressure on \((j,k)\).\footnote{If just those
  \(L^2\) features are used and the system is trained on a parallel corpus, the 
  value \(mn\lambda_{(m-1)L+n}\) may be considered as the ``effective
  weight'' of \(m\times n\) sub-segment pairs.}  Clearly, the
higher the pressure \(P_{jk}\), the higher the probability \(p(j,k)\)
is. In the absence of sub-segment information for any of the
\((j,k)\)'s of a particular segment pair, all probabilities are equal:
\(p(j,k)=\frac{1}{(|S||T|+|S|+|T|)}\). The \emph{pressures} are 
zero when either \(j\) or \(k\) is NULL.

\begin{figure}[!h]
  \centering
  \includegraphics[scale=0.6 ]{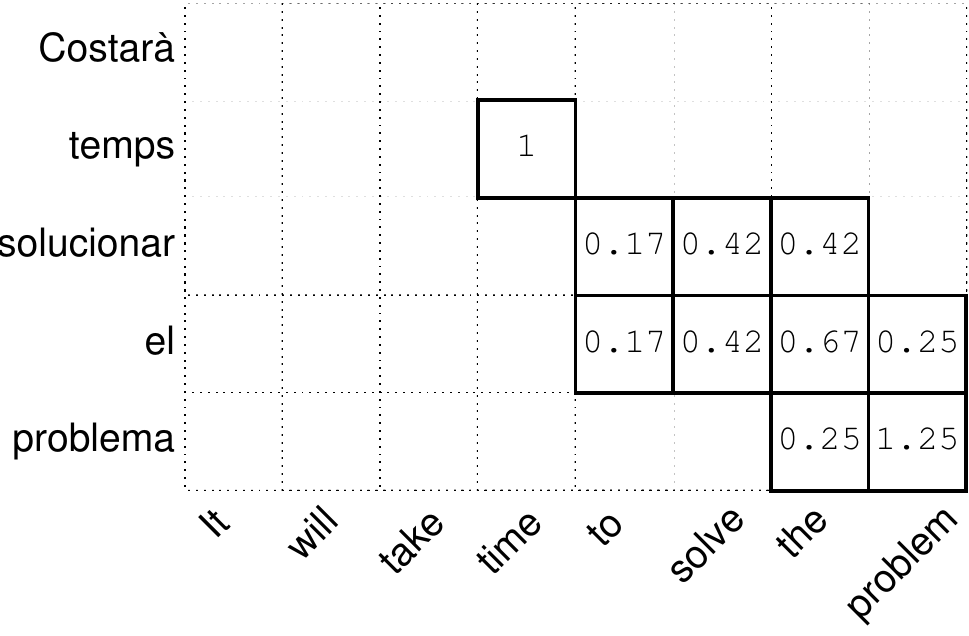}
  \caption{Alignment pressures.}
  \label{fig:alignmentstrength}
\end{figure}

Following our example, the alignment pressures for the words covered
by the sub-segment alignments are presented in Figure
\ref{fig:alignmentstrength}.  The word pair (\emph{temps},\emph{time})
is only covered by a sub-segment alignment (\emph{temps},
\emph{time}), so the surface is 1 and the alignment pressure is
$P_{2,4}=1$. On the other hand, the word pair (\emph{the},\emph{el}) is covered by
three sub-segment alignments: (\emph{solucionar el}, \emph{solve
  the}), (\emph{solucionar el}, \emph{to solve the}), and (\emph{el
  problema}, \emph{the problem}); therefore, the \emph{alignment pressure} is
$P_{4,7}=1/4+1/6+1/4=2/3\simeq 0.67$.

In this simple model, the alignment pressures \(P_{jk}\) themselves
may then be used instead of the probabilities \(p(j,k)\) to obtain
word-position alignments as described at the end of Section
\ref{se:alg-model}.





\begin{figure}
  \centering
  \includegraphics[scale=0.6]{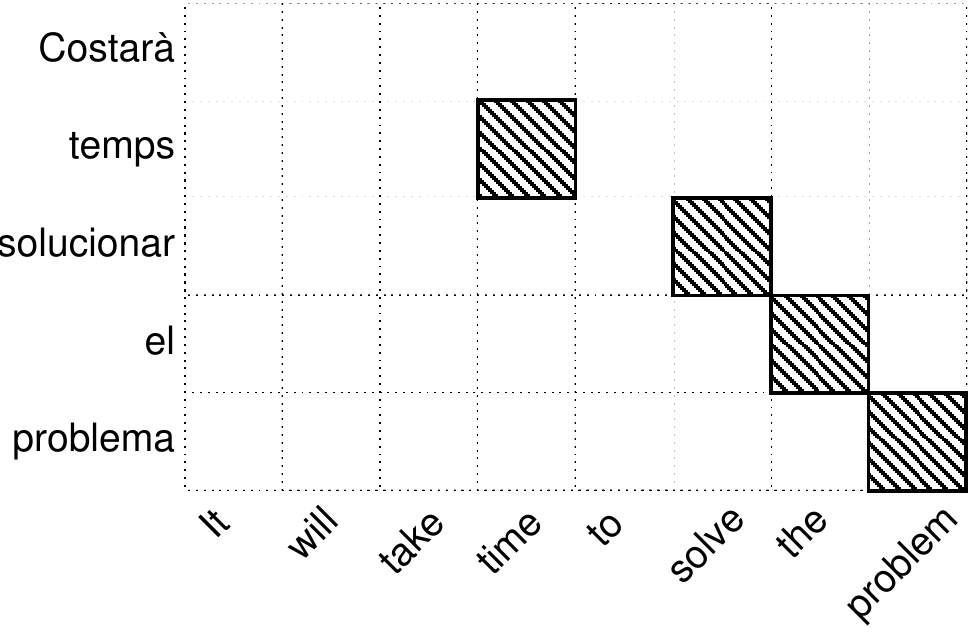}
  \includegraphics[scale=0.6]{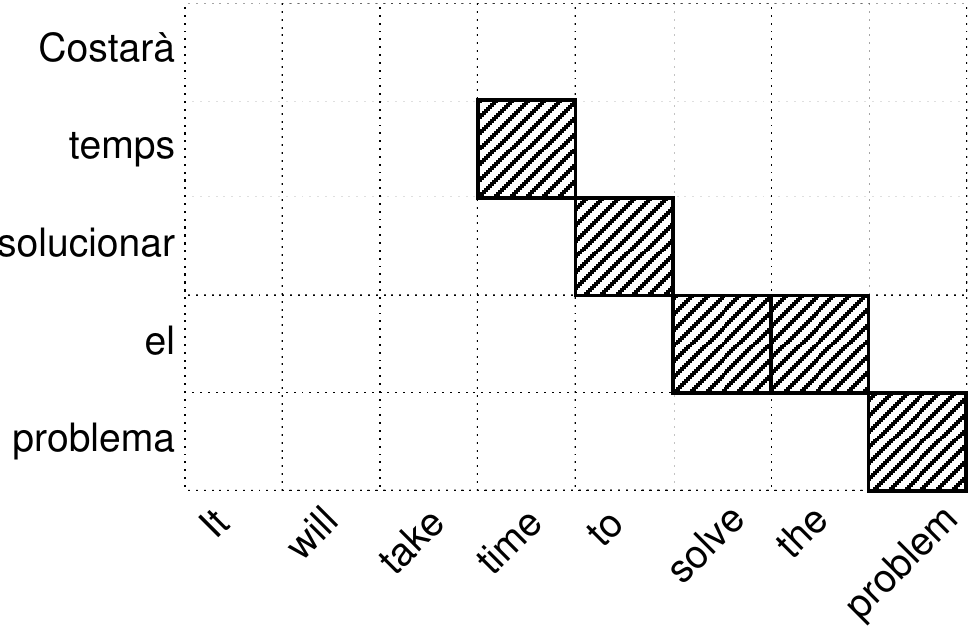}
  \caption{Resulting Catalan-to-English and English-to-Catalan word alignments.}
  \label{fig:st}
\end{figure}

As in the case of the general alignment model defined at the beginning
of this section, the alignment is performed both from source-to-target
and from target-to-source following the same procedure. Figure~\ref{fig:st}
shows the Catalan-to-English and the English-to-Catalan word alignments
for the running example. As can be seen, words \emph{to} and \emph{solve}
in English have the same alignment score for words \emph{solucionar} and
\emph{el} in Spanish, respectively. Therefore, the alignments closest
to the diagonal are chosen; in this case, \emph{to} is aligned
with \emph{solucionar}, and \emph{solve} is aligned with \emph{el} (not a very good alignment).
In the other direction of the alignment, the situation is similar
for word \emph{solucionar} in Spanish and words \emph{solve} and \emph{the}
in English (the resulting alignment is better here).

\begin{figure}
  \centering
  \includegraphics[scale=0.6]{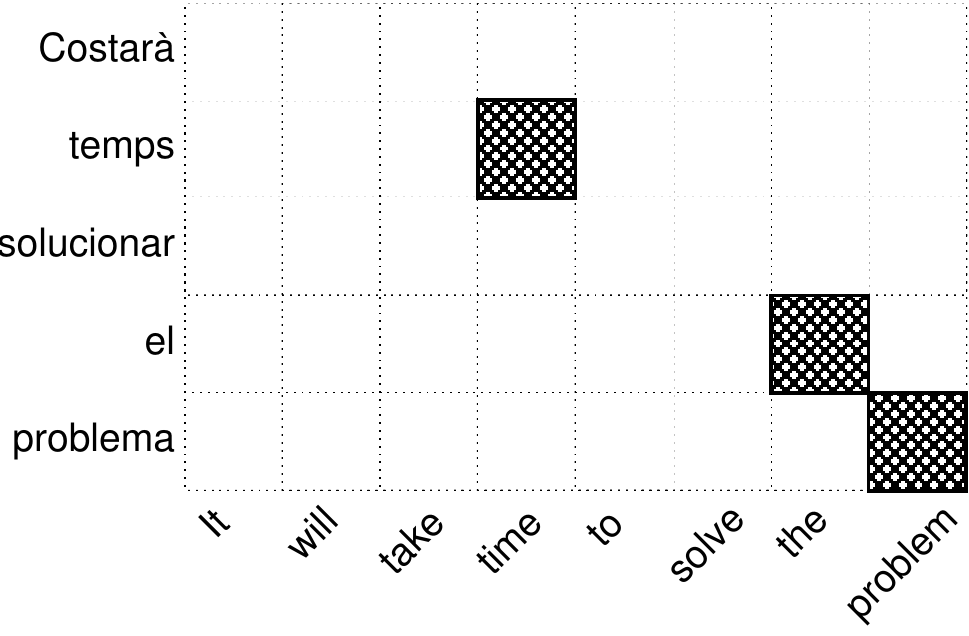}
  \includegraphics[scale=0.6]{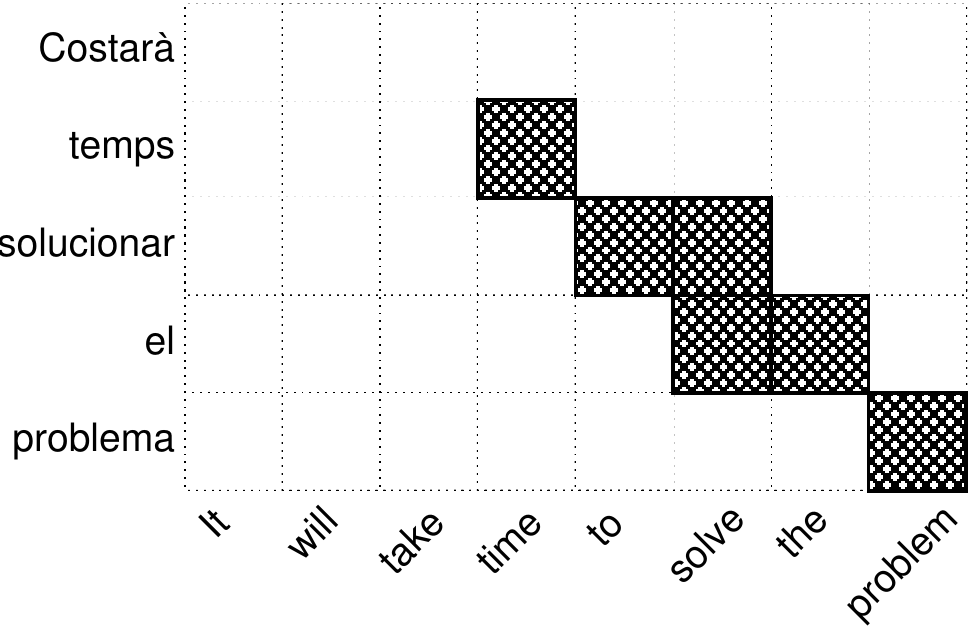}
  \caption{Two possible symmetrized word alignments, the first one using the intersection heuristic and the second one using the \emph{grow-diagonal-final-and} heuristic.}
  \label{fig:iters}
\end{figure}

Figure \ref{fig:iters} shows two possible symmetrized word alignments
obtained by computing, in the first case, the intersection
of the alignments shown in figure~\ref{fig:st},
and, in the second case, the the widely-used \emph{grow-diagonal-final-and}
heuristic of \cite{koehn03statis}, which, in this case, coincides
with the union of the alignments.

\section{Experiments} \label{se:exp}

In this section we describe the experimental setting designed for
measuring the performance of the alignment models described in
Section~\ref{se:alg-model}.  Two different experimental scenarios were
defined in order to measure (a) the quality of the alignments obtained
when using training corpora with several levels of reliability, and
(b) the domain independence of the weights trained for the parametric
aligner (P-aligner).

\paragraph{Gold-standard experiment.} For this experiment, we used the
EPPS \emph{gold
  standard}~\citep{springerlink:10.1007/s10579-005-4822-5}, a
collection of 500 pairs of sentences extracted from the
English--Spanish Europarl parallel corpus~\citep{koehn2005europarl} and
hand-aligned at the word level using two classes of alignments:
\emph{sure} alignments and \emph{possible} alignments.\footnote{Once
  the sub-segment alignments were obtained, the gold standard was
  lowercased to maximise the recall in the alignment process.} This
corpus was used for performing several evaluations:
\begin{itemize}
\item \emph{parametric alignment model (defined in
    Section~\ref{se:alg-model})}: we evaluated this model by using the
  gold standard corpus both for training and testing using a 10-fold
  cross-validation strategy. Therefore, for each fold we had 450 pairs
  of sentences as a training set and 50 pairs of sentences as a test
  set. We tried the two methods defined in Section~\ref{se:alg-model}
  for training: optimization of eq.~(\ref{eq:maximizeprobability}) by
  using a gradient ascent algorithm~\citep{duda2001pattern}, and
  minimizing directly the alignment error rate (AER) by using the
  \emph{simplex} algorithm~\citep{nelder1965simplex}. Increasingly
  large sets of bilingual sub-segments were used by defining different
  values of the maximum sub-segment length \(L\) in \([1,5]\).
 \item \emph{pressure aligner (defined in Section~\ref{ss:pressure})}: Since this
alignment model does not require training it was directly evaluated on the gold standard.
Increasingly large sets of bilingual sub-segments were used by defining different values
of the maximum sub-segment length \(L\) in \([1,5]\).
\item \emph{GIZA++ trained on the EPPS gold standard}:
  GIZA++~\citep{och03asc} was used as a baseline by repeating the
  previously described 10-fold cross-validation strategy.\footnote{The
    \emph{test-corpus} option in GIZA++ was used to train the
    alignment models with one corpus and then align another one.}
  Although it is obvious that 450 pairs of parallel sentences is not
  enough for obtaining high quality alignment models with this tool,
  this results are useful to measure the performance of the models
  proposed when using a very small training corpus.\footnote{To train GIZA++, the default configuration was used: 5
    iterations of model 1 and hidden Markov model and 3 iterations of
    models 3 and 4.}
\item \emph{GIZA++ trained on a large corpus}: In this experiment a
  larger corpus was used to train GIZA++ models: the English--Spanish
  parallel corpus provided for the machine translation task at the
  Seventh Workshop on Statistical Machine Translation (WMT12,
  \citet{callisonburch-EtAl:2012:WMT}), which includes the Europarl
  parallel corpus, from which the gold standard is extracted.  In this
  way, it is possible to compare the models proposed in this work with
  the use of the state-of-the-art tool GIZA++, which is commonly used
  in this scenario. This corpus is provided already aligned at the
  sentence level and, before training the alignment models, it was
  tokenised and lowercased, and sentences longer than 50 words were
  removed.\footnote{This preprocessing was performed by using the
    scripts provided by the Moses MT toolkit:
    \url{https://github.com/moses-smt/mosesdecoder/tree/master/scripts}[last visit:
     30th August 2012]}
\end{itemize}
Since all the alignment models proposed in this experiment are
asymmetric (i.e.  they must be trained from English to Spanish and
from Spanish to English separately) we experimented three different
symmetrization methods: intersection, union, and
\emph{grow-diagonal-final-and}~\citep{Axelrod05edinburghsystem}.

\paragraph{GIZA++ alignments as a reference.}

This second experiment focuses on measuring the re-usability of the
weights trained for the parametric alignment model. In this case, we
used three different corpora, all of them extracted from the
translation memory published by the \emph{European Commission
  Directorate-General for
  Translation}~\citep{dgttm2009}.\footnote{\url{http://langtech.jrc.it/DGT-TM.html}[last visit:
     30th August 2012]}
This translation memory is a collection of documents from the
\emph{Official Journal of the European
  Union}\footnote{\url{http://eur-lex.europa.eu}[last visit:
     30th August 2012]} which are provided
aligned at the sentence level. These documents are indexed by using a
set of domain
codes\footnote{\url{http://eur-lex.europa.eu/RECH_repertoire.do}[last visit:
     30th August 2012]}
which can be used to identify the documents belonging to the same
domain.  Following this method, we extracted three subsets from this
translation memory belonging to the domains: \emph{elimination of
  barriers to trade} (code 02.40.10.40), \emph{safety at work} (code
05.20.20.10), and \emph{general information of public contracts} (code
06.30.10.00). These corpora were chosen because they have similar
sizes (between 15894 and 13414 pairs of sentences) and they belong to
clearly different
domains, as evidenced by the cosine similarity measure\footnote{The cosine similarity was computed on the lowercased corpora, removing the punctuation signs and the stopwords defined in the Snowball project: \url{http://snowball.tartarus.org/algorithms/english/stop.txt}, \\
  \protect\url{http://snowball.tartarus.org/algorithms/spanish/stop.txt}[last visit:
     30th August 2012]}
presented in Table~\ref{fig:cosinus_sym}.\footnote{As a reference,
  note that if we split any of these three corpora into two parts and
  compute the cosine similarity between them, the results obtained are
  around 0.98.}


\begin{table}
 \centering
 \small{
  \begin{tabular}{ccc}
   \hline
   \multicolumn{1}{|c|}{\textbf{corpus en}} & \multicolumn{1}{c|}{\textbf{05.20.20.10}} & \multicolumn{1}{c|}{\textbf{06.30.10.00}} \\
   \hline
   \multicolumn{1}{|c|}{\textbf{02.40.10.40}} & \multicolumn{1}{c|}{0.21} & \multicolumn{1}{c|}{0.18} \\
   \hline
   \multicolumn{1}{|c|}{\textbf{06.30.10.00}} & \multicolumn{1}{c|}{0.15} & \\
   \cline{1-2}
  \end{tabular}
  \\[5mm]
  \begin{tabular}{ccc}
   \hline
   \multicolumn{1}{|c|}{\textbf{corpus es}} & \multicolumn{1}{c|}{\textbf{05.20.20.10}} & \multicolumn{1}{c|}{\textbf{06.30.10.00}} \\
   \hline
   \multicolumn{1}{|c|}{\textbf{02.40.10.40}} & \multicolumn{1}{c|}{0.22} & \multicolumn{1}{c|}{0.18} \\
   \hline
   \multicolumn{1}{|c|}{\textbf{06.30.10.00}} & \multicolumn{1}{c|}{0.13} & \\
   \cline{1-2}
  \end{tabular}
 \caption{Cosine similarity for both the English (en) and the Spanish (es) documents in the corpora released by 
   the \emph{European Commission Directorate-General for Translation} that we used.}

 \label{fig:cosinus_sym}
}
\end{table}

For this experiment, we followed these steps:
\begin{itemize}\itemsep 0ex
\item GIZA++ was used to align the three corpora and these alignments
  were taken as reference alignments;
\item using the three reference alignments as training corpora, three
  different sets of weights were obtained for the parametric aligner
  and each of these sets of weights was used to align the other two
  corpora and also the same corpus on which the weights were trained;
 \item the resulting alignments were compared with the reference alignments
to evaluate the re-usability of the weights in out-of-domain alignment tasks.
\end{itemize}

In addition, the GIZA++ alignment models obtained as a byproduct of
the computation of the reference alignments were also used to align
the test corpora.  We used the resulting alignments as a point of
comparison for the alignments produced by the parametric aligner.

The experiments were performed by using: a range of values for the
maximum sub-segment length \(L\), both the simplex and gradient ascent
algorithms for optimizing the weights of the parametric aligner, and
the three symmetrization methods previously commented. The best
results were obtained with \(L=5\) and the
\emph{grow-diagonal-final-and} symmetrization heuristic \citep{koehn03statis}.

\paragraph{Evaluation metrics.} 
For evaluating the different experiments defined in this section we
used the \emph{Lingua-AlignmentSet}
toolkit\footnote{\url{http://gps-tsc.upc.es/veu/personal/lambert/software/AlignmentSet.html}
  [last visit: 30th August 2012]} which computes, for a pair of
alignment set (\(A\)) and corresponding gold standard (\(G\)), the
precision (\(P\)), recall (\(R\)), and \(F\)-measure (\(F\))
\cite[Ch.\ 8.1]{manning1999foundations}, defined as usual:
\begin{displaymath}
P=|A \cap G|/|A| \hspace{2cm} R=|A \cap G|/|G| \hspace{2cm} F=2PR/(P+R)
\end{displaymath}
These measures are computed (a) only for the \emph{sure} alignments and (b) both for
\emph{sure} and \emph{possible} alignments. In addition, the alignment error rate (AER) is
computed by combining sure and possible alignments in the following way:
\begin{displaymath}
AER=1-\frac{|A \cap G_{\mathrm{sure}}|+|A \cap G|}{|A|+|G_{\mathrm{sure}}|}.
\end{displaymath}

\paragraph{Sources of bilingual information.} We used three different
machine translation (MT) systems to translate the sub-segments from
English into Spanish and vice versa, in order to get the sub-segment
alignments needed to obtain the features for the models defined in
Section~\ref{se:alg-model}:
\begin{itemize}\itemsep 0ex
 \item \emph{Apertium}:\footnote{\url{http://www.apertium.org} [last visit:
     30th August 2012]} a free/open-source platform for the development of
   rule-based MT systems~\citep{apertium2011}. We used the English--Spanish MT
   system from the project's
   repository\footnote{\url{https://apertium.svn.sourceforge.net/svnroot/apertium/trunk/apertium-en-es/}
     [last visit: 30th August 2012]} (revision 34706).
 \item \emph{Google Translate}:\footnote{\url{http://translate.google.com} [last visit: 30th August 2012]} an online MT system by Google Inc. (translations performed in July 2012).
 \item \emph{Microsoft Translator}:\footnote{\url{http://www.microsofttranslator.com} [last visit: 30th August 2012]} an online MT system by Microsoft (translations performed in July 2012).
\end{itemize}
It is worth noting that the Apertium system is oriented to
closely-related pairs of languages; furthermore, the Spanish--English
language pair is not as mature as other pairs in Apertium; therefore,
it is expected to produce translations of lower quality compared with
other state-of-the-art systems as indicated by observed BLEU
scores. For the gold-standard experiment, these three MT systems were
used. For the experiments using the translation memories released by
the \emph{European Commission Directorate-General for Translation},
only Apertium and Google could be used, given the huge amount of
sub-segments to be translated and the restrictions in the Microsoft
Translator API.

\section{Results and discussion} \label{se:results}

This section presents the results obtained in the experiments
described in the Section~\ref{se:exp}. Table~\ref{tb:results1} shows
the results in terms of precision (\(P\)), recall (\(R\)),
\(F\)-measure (\(F\)) and alignment error rate (AER) obtained by both
the parametric aligner (P-aligner) described in
Section~\ref{se:alg-model}, the ``pressure'' aligner described in
Section~\ref{ss:pressure}, and GIZA++ both when using a 10-fold
cross-validation strategy on the gold standard corpus and when using
the corpus from the WMT12 workshop for training the alignment
models. It is worth noting that the results computed using the 10-fold
cross-validation (\emph{P-aligner probability optimization},
\emph{P-aligner AER optimization}, and \emph{GIZA++ trained on the
  gold standard}) are presented as the average of the results obtained
in each fold. The parametric aligner was both trained by using all the
alignments available in the training sets and only using the sure
ones. The results of the parametric aligner (best AER in the
27\%--29\% range) overcame, as expected, the results obtained by the
``pressure'' aligner (AER around 32\%), since the weights were trained
on a gold standard and not fixed beforehand.\footnote{The results of
  the ``pressure'' aligner come however surprisingly close.} As can be
appreciated, both the P-aligner and the ``pressure'' aligner overcame
the results by GIZA++ trained on the gold standard for all the metrics
used (AER around 55\%). This is easily explainable given the small
size of the corpus used to train the alignment models with GIZA++. In
any case, this shows the convenience of our model when using a very
reduced training corpus. Finally, the alignments from GIZA++ trained
on the WMT12 corpus obtained the best results in terms of F-measure
and AER (16\%). If precision and recall are compared, one can see that
the precision in both GIZA++ and the parametric aligner are quite
similar but GIZA++ obtains better results in recall.  This is an
interesting result, since this means that, for tasks like CAT
\citep{espla11b}, where precision is more relevant than the
recall, the parametric aligner may be as useful as GIZA++. Also, this
means that using more (or better) sources of bilingual information
could help to obtain closer results to those obtained by GIZA++ in
recall and, consequently, in F-measure and AER. To
understand these results better, a complementary experiment was performed
by using several sub-sets from the WMT12 corpus with different
sizes. We found out that, to obtain the same results produced by
the P-aligner in terms of AER, GIZA++ requires an in-domain training
corpus with a size between 5,000 pairs of sentences (AER 29.5\%) and
10,000 pairs of sentences (AER 26.2\%). This confirms that GIZA++
requires a considerably larger training corpus than that needed by the
proposed approach and, as a consequence, it would be quite difficult
to use it for aligning sentences on the fly or for small amounts of
corpora.

There are some differences in the results obtained for the P-aligner
depending on the training method used: the model trained through the
maximization of the total alignment probability obtained higher
results in precision (91\% versus 75\%), whereas the model trained by
minimizing the AER provided better results for recall (65\% versus
56\%). Although the results for F-measure and AER are very similar,
they happen to be slightly better when using the minimization of the
AER, as expected, since in this case the evaluation function is
directly optimized during the training process.

\bgroup
\setlength{\tabcolsep}{0.45em}
\begin{table}[ht]
\centering
\footnotesize{
  \begin{tabular}{|c|c||ccc||ccc||c|}
   \cline{2-9}
   \multicolumn{1}{c|}{} & \(L\) & \textbf{\(P_s\)} & \textbf{\(R_s\)} & \textbf{\(F_s\)} & \textbf{\(P\)} & \textbf{\(R\)} & \textbf{\(F\)} & \textbf{\(\mathrm{AER}\)} \\
   \hline
   \multicolumn{2}{|c||}{\textbf{GIZA++ trained on the gold standard}} & 48.0\% & 40.0\% & 43.6\% & 52.2\% & 30.4\% & 38.4\% & 54.5\% \\
   \hline
   \hline
   \multicolumn{2}{|c||}{\textbf{GIZA++ 5,000 sentences of WMT12 corpus}} & 66.6\% & 66.2\% & 66.4\% & 74.7\% & 52.0\% & 61.3\% & 29.5\% \\
   \hline
   \hline
   \multirow{5}{*}{\textbf{P-aligner probability optimization}} & 1 & 86.0\% & 44.2\% & 58.3\% & 89.9\% & 32.4\% & 47.6\% & 40.3\% \\
   \cline{2-9}
   & 2 & 88.3\% & 52.9\% & 66.1\% & 92.2\% & 38.7\% & 54.5\% & 32.5\% \\
   \cline{2-9}
   & 3 & 90.1\% & 55.7\% & 68.8\% & 94.0\% & 40.7\% & 56.8\% & 29.7\% \\
   \cline{2-9}
   & 4 & 91.0\% & 56.4\% & 69.6\% & 94.9\% & 41.2\% & 57.4\% & 28.9\% \\
   \cline{2-9}
   & 5 & 91.4\% & 56.2\% & 69.6\% & 95.2\% & 41.1\% & 57.3\% & 29.0\% \\
   \hline
   \hline
   \multirow{5}{*}{\textbf{P-aligner AER optimization}} & 1 & 81.6\% & 52.5\% & 63.9\% & 85.6\% & 38.6\% & 53.2\% & 34.6\% \\
   \cline{2-9}
   & 2 & 71.7\% & 60.0\% & 65.3\% & 78.7\% & 46.1\% & 58.1\% & 31.5\% \\
   \cline{2-9}
   & 3 & 73.7\% & 63.8\% & 68.4\% & 81.4\% & 49.5\% & 61.4\% & 28.1\% \\
   \cline{2-9}
   & 4 & 75.3\% & 64.5\% & 69.5\% & 82.7\% & 49.7\% & 62.0\% & 27.1\% \\
   \cline{2-9}
   & 5 & 74.8\% & 65.4\% & 69.8\% & 82.4\% & 50.6\% & 62.6\% & 26.7\% \\
   \hline
   \hline
   \multirow{5}{*}{\textbf{``pressure'' aligner}} & 1 & 80.4\% & 39.1\% & 52.6\% & 85.0\% & 28.9\% & 43.1\% & 45.9\% \\
   \cline{2-9}
   & 2 & 70.9\% & 54.0\% & 61.3\% & 76.9\% & 40.9\% & 53.4\% & 36.1\% \\
   \cline{2-9}
   & 3 & 69.8\% & 58.0\% & 63.3\% & 76.7\% & 44.5\% & 56.3\% & 33.6\% \\
   \cline{2-9}
   & 4 & 69.2\% & 59.0\% & 63.7\% & 76.3\% & 45.5\% & 57.0\% & 33.0\% \\
   \cline{2-9}
   & 5 & 69.1\% & 59.4\% & 63.9\% & 76.3\% & 45.8\% & 57.3\% & 32.8\% \\
   \hline
   \hline
   \multicolumn{2}{|c||}{\textbf{GIZA++ 10,000 sentences of WMT12 corpus}} & 69.2\% & 69.8\% & 69.5\% & 77.7\% & 54.8\% & 64.3\% & 26.2\% \\
   \hline
   \hline
   \multicolumn{2}{|c||}{\textbf{GIZA++ trained on whole WMT12}} & 77.2\% & 80.6\% & 78.9\% & 87.3\% & 63.7\% & 73.7\% & 16.0\% \\
   \hline
  \end{tabular}
}
\caption{Average values of precision (\(P\)), recall (\(R\)), \(F\)-measure
  (\(F\)), and alignment error rate (AER) for 
  the alignments obtained with GIZA++ (when trained both on the gold standard and several portions of the WMT12 parallel corpus),
  and the parametric aligner (P-aligner) trained by optimizing the total alignment probabilities,
  and by optimizing the AER, for different values of the maximum sub-segment length \(L\). The results obtained by the ``pressure'' aligner are also reported. 
  The training of the parametric aligner was performed by using only the sure alignments.}
 \label{tb:results1}
\end{table}
\egroup


Finally, Table~\ref{tb:results-indepdenence} shows the results obtained
for the experiment with the translation memories from the \emph{Official
Journal of the European Union}, which is aimed at measuring the domain-independence
of the weights trained for the parametric aligner. The table shows, for the parametric aligner (using
both training methods) and GIZA++, the results obtained when training on one of the
corpora and aligning the other two corpora. The results reported in this table
were obtained by using sub-segments of length \(L=5\), as this setting  provided the
best results. As in the previous experiments, the symmetrization technique used
was \emph{grow-diagonal-final-and} \citep{och03asc}. As can be seen, the results
for all the parametric aligners compared are quite similar for all the systems
and all the training/test corpora (AER in the range 27\%--34\%). 
It is worth mentioning
that in this particular experiment the alignments produced by GIZA++ are being used as a gold
standard for evaluation, which could be unfair for our system, since some correct alignments
from the P-aligner could be judged as incorrect. Nevertheless, when the corpora
used for testing is different from that used for evaluation, the parametric aligners
obtain better results than GIZA++ (AER in the range 30\%--40\%), but the most important
finding is the relative uniformity in the results when using different corpora for
training and aligning. This shows that the weights learned from a corpus in a given
domain can be re-used to align corpora in different domains. This is a very desirable
property, as it would imply that, in a real application, 
once the aligner is trained, it can be used for aligning
any new pair of sentences \emph{on the fly}.

\bgroup
\setlength{\tabcolsep}{0.45em}
\begin{table}[ht]
 \centering
\footnotesize{
  \begin{tabular}{|c|c|c||cccc|}
   \cline{2-7}
   \multicolumn{1}{c|}{} & \textbf{training} & \textbf{test} & \textbf{\(P\)} & \textbf{\(R\)} & \textbf{\(F\)} & \textbf{\(\mathrm{AER}\)} \\
   \hline
   \multirow{9}{*}{\textbf{P-aligner probability optimization}} & \multirow{3}{*}{02.04.10.40} & 02.04.10.40 & 73.12\% & 66.61\% & 69.71\% & 30.29\% \\
   \cline{3-7}
   & & 05.20.20.10 & 79.3\% & 68.4\% & 73.4\% & 26.6\% \\
   \cline{3-7}
   & & 06.30.10.00 & 77.0\% & 65.5\% & 70.8\% & 29.3\% \\
   \cline{2-7}
   \cline{2-7}
   & \multirow{3}{*}{05.20.20.10} & 02.04.10.40 & 72.8\% & 64.2\% & 68.2\% & 31.8\% \\
   \cline{3-7}
   & & 05.20.20.10 & 79.61\% & 66.29\% & 72.34\% & 27.66\% \\
   \cline{3-7}
   & & 06.30.10.00 & 78.2\% & 63.6\% & 70.1\% & 29.9\% \\
   \cline{2-7}
   \cline{2-7}
   & \multirow{3}{*}{06.30.10.00} & 02.04.10.40 & 71.9\% & 63.9\% & 67.7\% & 32.4\% \\
   \cline{3-7}
   & & 05.20.20.10 & 78.6\% & 65.5\% & 71.5\% & 28.5\% \\
   \cline{3-7}
   & & 06.30.10.00 & 77.5\% & 63.2\% & 69.6\% & 30.4\% \\
   \hline
   \hline
   \multirow{9}{*}{\textbf{P-aligner AER optimization}} & \multirow{3}{*}{02.04.10.40} & 02.04.10.40 & 73,1\% & 60,3\% & 66,1\% & 33,9\% \\
   \cline{3-7}
   & & 05.20.20.10 & 80.5\% & 65.5\% & 72.3\% & 27.8\% \\
   \cline{3-7}
   & & 06.30.10.00 & 78.3\% & 63.0\% & 69.8\% & 30.2\% \\
   \cline{2-7}
   \cline{2-7}
   & \multirow{3}{*}{05.20.20.10} & 02.04.10.40 & 71.2\% & 64.6\% & 67.8\% & 32.3\% \\
   \cline{3-7}
   & & 05.20.20.10 & 79,7\% & 67,4\% & 73,1\% & 26,9\% \\
   \cline{3-7}
   & & 06.30.10.00 & 76.1\% & 63.0\% & 68.9\% & 31.1\% \\
   \cline{2-7}
   \cline{2-7}
   & \multirow{3}{*}{06.30.10.00} & 02.04.10.40 & 70.5\% & 68.8\% & 69.6\% & 30.4\% \\
   \cline{3-7}
   & & 05.20.20.10 & 75.6\% & 70.2\% & 72.8\% & 27.2\% \\
   \cline{3-7}
   & & 06.30.10.00 & 74,6\% & 67,4\% & 70,8\% & 29,2\% \\
   \hline
   \hline
   \multirow{9}{*}{\textbf{GIZA++}} & \multirow{3}{*}{02.04.10.40} & 02.04.10.40 & 83.2\% & 81.7\% & 82.5\% & 17.5\% \\
   \cline{3-7}
   & & 05.20.20.10 & 71.3\% & 64.3\% & 67.6\% & 32.4\% \\
   \cline{3-7}
   & & 06.30.10.00 & 67.5\% & 62.4\% & 64.9\% & 35.1\% \\
   \cline{2-7}
   & \multirow{3}{*}{05.20.20.10} & 02.04.10.40 & 70.9\% & 61.9\% & 66.1\% & 33.9\% \\
   \cline{3-7}
   & & 05.20.20.10 & 90.0\% & 89.6\% & 89.8\% & 10.2\% \\
   \cline{3-7}
   & & 06.30.10.00 & 72.9\% & 68.0\% & 70.3\% & 29.7\% \\
   \cline{2-7}
   & \multirow{3}{*}{06.30.10.00} & 02.04.10.40 & 64.1\% & 55.8\% & 59.7\% & 40.4\% \\
   \cline{3-7}
   & & 05.20.20.10 & 70.2\% & 63.6\% & 66.8\% & 33.2\% \\
   \cline{3-7}
   & & 06.30.10.00 & 87.4\% & 87.4\% & 87.4\% & 12.6\% \\
   \hline
   \hline
  \end{tabular}
  \caption{Precision (\(P\)), recall (\(R\)), \(F\)-measure (\(F\)), and
    alignment error rate (AER) for the alignments obtained with 
    the parametric   aligner (P-aligner) trained by optimizing the
    total alignment probabilities, the P-aligner trained by optimizing the AER, 
    and  GIZA++ when using corpora from different domains for
    training and testing. }
 \label{tb:results-indepdenence}
}
\end{table}

\section*{Concluding remarks and future work}
In this work we have described a new approach for word alignment based
on the use of sources of bilingual information that makes no
assumptions about the languages of texts being aligned. Two alignment
methods have been proposed: (a) an intuitive and training-free aligner
based on the idea of the pressure exerted on the word-pair squares of
a sentence-pair rectangular grid by the bilingual sub-segments
(rectangles) covering words in both sentences to be aligned, and (b) a
more general maximum-entropy-style (``log-linear'') parametric aligner
which may be seen as a generalization of that aligner. A set of
experiments was performed to evaluate both approaches, comparing them
with the state-of-the-art tool GIZA++. The results obtained show that
the models proposed obtain results comparable to those obtained by the
state-of-the-art tools in terms of precision. Although GIZA++ obtains
better results in recall and in general measures, such as
\(F\)-measure and AER (16\%), the parametric aligner overcomes GIZA++
(AER 54\%) when using a small training corpus. In addition, the
results show that the weights trained for the parametric aligner can
be re-used to align sentences from different domains to the one from
which they were trained. In this case the new approach provides better
results than GIZA++ when aligning out-of-domain corpora. This means
that it is possible to use the proposed alignment models to align new
sentences \emph{on the fly}, which can be specially useful in some
scenarios as the case of computer-aided translation (CAT).

As a future work, we plan to perform wider experiments including other
pairs of languages and also other sources of bilingual
information. Note that the parameters of the parametric MT-based
aligner proposed here could also be intrinsically optimized according
to the overall performance of a larger task using alignment as a
component, such as \emph{phrase}-based SMT.

\paragraph{Acknowledgements:} Work partially supported by the Spanish
Ministry of Science and Innovation through project TIN2009-14009-C02-01,
and by the Universitat d'Alacant through project GRE11-20.



\bibliographystyle{apalike} 
\bibliography{onthefly}
\end{document}